%% file: main_ISBI2027.tex
\documentclass{article}
\usepackage{spconf}
\usepackage{graphicx}
\usepackage{xcolor}
\usepackage{amsmath,amssymb}
\usepackage{booktabs}
\usepackage{multirow}
\usepackage{cite}
\usepackage{balance}
\usepackage[colorlinks]{hyperref}
\let\originalCaronAccent\v

\input{defs}

\graphicspath{{Figs/}}
\newcommand{\msd}[2]{#1{\scriptsize$\pm$#2}}
\newcommand{\bmsd}[2]{\textbf{#1}{\scriptsize$\boldsymbol{\pm}$\textbf{#2}}}

\title{%Topology-Informed Prompt-Conditioned Universal Segmentation for Gynecologic Ultrasound and MRI
Topology-Informed Prompt-Conditioned Universal Segmentation of Uterine Structures from Ultrasound and MRI}
\name{Yongheng~Sun$^{1}$, 
Yuexi Gu$^{1}$, 
Jingwen Sun$^{1}$, 
Maureen Kohi$^{1,*}$, 
Mingxia~Liu$^{1,*}$}
\address{
$^1$Department of Radiology and BRIC, UNC at Chapel Hill, Chapel Hill, NC 27599, USA}
\begin{document}
\ninept
\maketitle
\begin{abstract}
Multi-structure segmentation of the uterus is important for computer-assisted screening, diagnosis, and treatment planning of uterine diseases, where ultrasound and MRI provide complementary clinical information. 
However, developing a unified model across these modalities is challenging due to their substantially different image appearances, anatomical contexts, spatial resolutions, and label spaces. 
Moreover, existing datasets often define different segmentation targets, making joint learning challenging and potentially leading to negative transfer across heterogeneous tasks. 
To this end, we propose a Topology-informed Prompt-conditioned Universal Segmentation ({TPUS}) framework for 
{%gynecologic ultrasound and MRI}
segmenting multiple uterine structures across ultrasound and MRI}.  
TPUS introduces a graph-based multi-dataset backbone comprising modality-specific stems and a modality-shared graph-based encoder-decoder to support modality-sensitive input adaptation, structural feature reasoning, and joint representation learning across heterogeneous uterine segmentation tasks. 
In addition,  TPUS uses task-aware class prompts to condition the segmentation process for different datasets and label spaces, a dynamic convolutional adaptation module to generate task-specific output responses, and a topology-informed loss to encourage anatomically consistent predictions. 
Experiments on a uterine ultrasound dataset and a T2-weighted uterine myoma MRI dataset demonstrate that TPUS achieves Dice scores of 0.898 and 0.693 on the two held-out test sets, respectively, outperforming several generic and universal segmentation baselines. 
Source code can be accessed \href{https://github.com/YonghengSun1997/TPUS}{online}.
%and highlight its potential in advancing automated radiology report generation.
\end{abstract}

\section{INTRODUCTION}
\label{sec:intro}

Multi-structure segmentation of the uterus is an important task in medical image analysis, supporting computer-assisted screening, diagnosis, disease characterization, and treatment planning for a wide range of uterine diseases~\cite{bones2024automatic,pan2024umd}. 
Accurate delineation of the uterus and its internal structures, such as the uterine cavity and myometrium, as well as focal lesions including uterine fibroids and Nabothian cysts, can provide quantitative anatomical and pathological information for clinical assessment~\cite{pan2024umd}. 
Ultrasound and MRI are two commonly used imaging modalities for evaluating uterine anatomy and pathology~\cite{bones2024automatic,pan2024umd}, offering complementary information regarding accessibility, spatial detail, soft-tissue contrast, and tissue characterization. 
Developing robust methods to segment multiple uterine structures and lesions across ultrasound and MRI could facilitate comprehensive automated analysis of uterine diseases.

However, achieving unified segmentation across ultrasound and MRI remains challenging due to substantial differences between the two modalities in image appearance, anatomical context, spatial resolution, and tissue contrast~\cite{bones2024automatic,pan2024umd}. 
Beyond these modality-specific differences, publicly available datasets are often designed for different clinical objectives and therefore provide heterogeneous segmentation targets and label definitions. 
For example, one dataset may focus on delineating the uterus and its internal structures, whereas another may primarily target uterine lesions such as fibroids. Consequently, directly combining such datasets for joint model training can introduce substantial task heterogeneity and lead to negative transfer, where knowledge beneficial to one segmentation task may interfere with another. 
Most existing methods are designed for a single modality, dataset, or fixed label space, and thus are not well suited for jointly learning heterogeneous uterine segmentation tasks across ultrasound and MRI.

\begin{figure*}[!t]
\setlength{\abovecaptionskip}{-8pt}
\setlength{\belowcaptionskip}{0pt}
\setlength{\abovedisplayskip}{0pt}
\setlength\belowdisplayskip{0pt}
    \centering  \includegraphics[width=\linewidth]{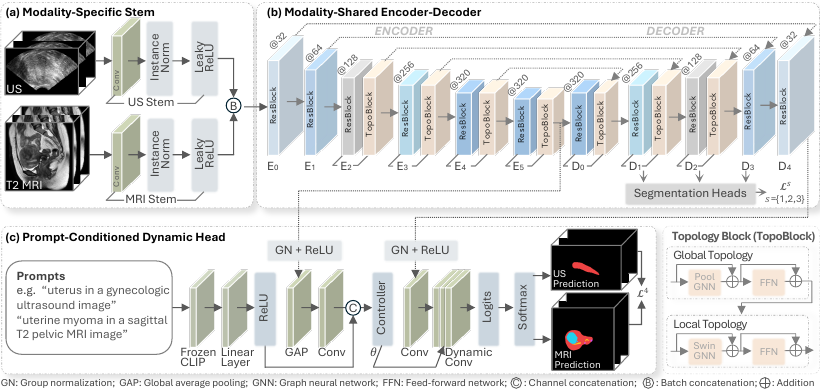} % 图片路径
    \if false 
    \caption{Overview of the proposed TPUS framework. (a) A sampled 3D ultrasound (US) or T2-weighted MRI volume is processed by its corresponding modality-specific stem. (b) The modality-shared %GNN-enhanced 
    encoder-decoder extracts multi-scale features, with heads at three decoders ($\mathrm{D}_1$--$\mathrm{D}_3$) providing auxiliary predictions. (c) Task-aware class prompts encoded by the frozen CLIP text encoder and the encoder ($\mathrm{E}_5$) bottleneck condition a class-wise dynamic head applied to the decoder ($\mathrm{D}_4$) for full-resolution prediction.}
    \fi 
    \caption{Overview of the proposed TPUS framework. 
(a) Modality-specific stems process input 3D US or MRI volumes. 
(b) The modality-shared graph-based encoder-decoder with  
topology blocks ({TopoBlocks}) extracts multi-scale features, with auxiliary heads at decoders $\mathrm{D}_1$--$\mathrm{D}_3$. 
(c) Task-aware class prompts from CLIP~\cite{radford2021learning} and the $\mathrm{E}_5$ bottleneck condition a dynamic head at $\mathrm{D}_4$ for full-resolution prediction.}
\label{fig_pipeline}
\end{figure*}

To address these challenges, we propose a
Topology-informed Prompt-conditioned Universal Segmentation (\textbf{TPUS}) framework for unified multi-structure segmentation across uterine ultrasound and MRI. 
As shown in Fig.~\ref{fig_pipeline}, TPUS is designed to jointly learn representations across heterogeneous datasets while adapting the segmentation process to dataset-specific label spaces. 
Specifically, we introduce task-aware class prompts to explicitly condition the model on different segmentation targets, enabling flexible prediction across heterogeneous label definitions. 
A dynamic convolutional adaptation module further generates task-specific output responses, allowing the representation to be effectively adapted to different modalities and segmentation tasks. 
To preserve anatomical consistency, we incorporate a topology-informed loss that encourages structurally plausible predictions. 
In addition, we develop a graph-based multi-dataset backbone comprising modality-specific stems and a modality-shared graph-based encoder-decoder, enabling modality-sensitive input adaptation and joint representation learning across heterogeneous uterine segmentation tasks.
These components enable TPUS to jointly learn from ultrasound and MRI while mitigating negative transfer and improving multi-structure uterine segmentation.
Experiments on uterine ultrasound and T2-weighted MRI datasets demonstrate that the proposed TPUS consistently outperforms several generic and universal segmentation methods.

\section{METHODOLOGY}
\label{sec:method}
{The proposed TPUS addresses cross-modality appearance variation and dataset-specific label spaces through modality-sensitive input adaptation, joint representation learning, prompt-conditioned prediction, and topology-informed deep supervision (Fig.~\ref{fig_pipeline}).} 
%方法一开始上来得有个说法吧？
%之前我们详细说过怎么写文章 回忆一下

% Motivation最重要 为什么这么做 ；现在写的是个实验报告

\subsection{Graph-Based Multi-Dataset Backbone}
The graph-based multi-dataset backbone comprises modality-specific stems and a modality-shared graph-based encoder-decoder. 
Let $\mathcal{D}=\{\mathcal{D}^{(d)}\}_{d=1}^{K}$ denote $K$ datasets, where dataset $d$ provides an image $\mathbf{x}^{(d)}$ and a label map $\mathbf{y}^{(d)}$ defined over its own class set $\mathcal{C}_d$.
TPUS follows the MultiTalent multi-dataset paradigm~\cite{ulrich2023multitalent}, while separating modality-sensitive input processing from shared anatomical representation learning.
As shown in Fig.~\ref{fig_pipeline}(a), each dataset has an independent stem $\mathcal{S}_d$ comprising a $3\!\times\!3\!\times\!3$ convolution, instance normalization, and LeakyReLU that maps its one-channel volume to 32 channels:
\begin{equation}
    \mathbf{z}^{(d)}_0=\mathcal{S}_d\!\left(\mathbf{x}^{(d)}\right).
    \label{eq:stem}
\end{equation}
At each iteration, one sampled volume is routed through its corresponding stem. With the reported batch size of one, the concatenation operator in Fig.~\ref{fig_pipeline}(a) simply forwards the selected stem output rather than pairing US and MRI volumes. The output then enters the modality-shared six-level residual U-shaped encoder-decoder in Fig.~\ref{fig_pipeline}(b), whose encoder widths are 32, 64, 128, 256, 320, and 320 channels and whose decoder widths are 320, 256, 128, 64, and 32 channels. Resolution-matched skip connections link the two paths; all parameters after the stems are shared.

Within the modality-shared graph-based encoder-decoder, we adapt the Pool graph neural network (GNN) and Swin GNN~\cite{shi2023nextou} into several topology blocks ({TopoBlocks}) deployed at encoder levels $\mathrm{E}_2$--$\mathrm{E}_5$ and decoder levels $\mathrm{D}_0$--$\mathrm{D}_2$; high-resolution levels $\mathrm{E}_0$--$\mathrm{E}_1$ and $\mathrm{D}_3$--$\mathrm{D}_4$ retain convolutional residual blocks. 
The TopoBlock shown in the bottom right of Fig.~\ref{fig_pipeline} applies graph reasoning and feed-forward network (FFN) refinement residually to feature $\mathbf{F}$:
\begin{align}
    \widehat{\mathbf{F}}_{p} &= \mathcal{G}_{p}(\mathbf{F})+\mathbf{F}, &
    \mathbf{F}_{p} &= \operatorname{FFN}(\widehat{\mathbf{F}}_{p})+\widehat{\mathbf{F}}_{p}, \nonumber\\
    \widehat{\mathbf{F}}_{w} &= \mathcal{G}_{w}(\mathbf{F}_{p})+\mathbf{F}_{p}, &
    \mathbf{F}' &= \operatorname{FFN}(\widehat{\mathbf{F}}_{w})+\widehat{\mathbf{F}}_{w},
    \label{eq:gnn}
\end{align}
where $\mathcal{G}_{p}$ denotes Pool GNN, which builds a dynamic $k$-nearest-neighbor graph after max pooling and restores resolution by max unpooling, and $\mathcal{G}_{w}$ denotes Swin GNN, which builds dynamic graphs within shifted 3D windows. Both use max-relative graph convolution to complement the backbone's local convolutional bias.

% Preserve first-appearance citation order despite the early table declaration.

\subsection{Prompt-Conditioned Dynamic Head}
Figure~\ref{fig_pipeline}(c) shows the class-text, image-context, and prediction paths of the dynamic head. For each class $c\in\mathcal{C}_d$, including background, we form a task-aware prompt, such as ``uterus in a gynecologic ultrasound image'' or ``uterine myoma in a sagittal T2 pelvic MRI image.'' A frozen CLIP ViT-B/32 text encoder~\cite{radford2021learning} yields an embedding $\mathbf{e}_{d,c}\in\mathbb{R}^{512}$. Then TPUS combines class semantics with image context to generate the classifier. The $\mathrm{E}_5$ bottleneck feature $\mathbf{B}$ and class embedding are projected to 256 dimensions:
\begin{align}
    \mathbf{i} &= \operatorname{Conv}\!\left(\operatorname{GAP}
    \left(\operatorname{ReLU}(\operatorname{GN}(\mathbf{B}))\right)\right), \nonumber\\
    \mathbf{t}_{d,c} &= \operatorname{ReLU}(\mathbf{W}_{t}\mathbf{e}_{d,c}+\mathbf{b}_{t}), \nonumber\\
    \boldsymbol{\theta}_{d,c} &= \mathcal{H}\!\left([\mathbf{i}\,\|\,\mathbf{t}_{d,c}]\right),
    \label{eq:conditioning}
\end{align}
where $[\cdot\|\cdot]$ denotes channel-wise concatenation at the circled C in Fig.~\ref{fig_pipeline}(c),  the shared $1\!\times\!1\!\times\!1$ controller $\mathcal{H}$ maps the 512-D condition to $\boldsymbol{\theta}_{d,c}\in\mathbb{R}^{153}$, GAP denotes global average pooling, ReLU represents the rectified linear unit, and GN represents group normalization.

In the prediction path, the final $\mathrm{D}_4$ decoder feature $\mathbf{F}_{4}\in\mathbb{R}^{32\times D\times H\times W}$ is transformed into an eight-channel map:
\begin{equation*}
    \mathbf{Q}=\operatorname{Conv}%_{1^3}
    \!\left(
    \operatorname{ReLU}(\operatorname{GN}(\mathbf{F}_{4}))\right). 
\end{equation*}
The dynamic convolution (DynConv) block in Fig.~\ref{fig_pipeline}(c) comprises three class-wise dynamic convolutions parameterized by $\boldsymbol{\theta}_{d,c}$:
\begin{equation}
    \begin{aligned}
        \mathbf{q}^{1}_{d,c} &= \sigma(\mathbf{W}^{1}_{d,c}*\mathbf{Q}+\mathbf{b}^{1}_{d,c}),\\
        \mathbf{q}^{2}_{d,c} &= \sigma(\mathbf{W}^{2}_{d,c}*\mathbf{q}^{1}_{d,c}+\mathbf{b}^{2}_{d,c}),\\
        \mathbf{o}_{d,c} &= \mathbf{W}^{3}_{d,c}*\mathbf{q}^{2}_{d,c}+\mathbf{b}^{3}_{d,c},
    \end{aligned}
    \label{eq:dynamic_head}
\end{equation}
where $*$ is a $1\!\times\!1\!\times\!1$ convolution, $\sigma$ is ReLU, and the channel dimensions are $8\!\rightarrow\!8\!\rightarrow\!8\!\rightarrow\!1$.
Each class therefore requires $(8^2+8)+(8^2+8)+(8+1)=153$ generated parameters. The controller is shared across datasets and classes, but the resulting filters depend on both image and class. CLIP embeddings remain fixed, whereas their projection and the controller are trainable. This head replaces the fixed $\mathrm{D}_4$ classifier and produces the native multiclass logits; softmax yields the corresponding prediction. Unlike previous CLIP-driven dynamic convolution~\cite{liu2023clipdriven}, which predicts a fixed foreground vocabulary using independent sigmoid outputs, TPUS uses task-aware class prompts to generate background-inclusive softmax logits in each dataset's native label space.

\subsection{Topology-Informed Deep Supervision}
% {\color{red}The loss symbols in Fig.~\ref{fig_pipeline}(b)--(c)} denote supervision of the auxiliary decoder heads and the full-resolution dynamic output. 
We apply deep supervision to outputs at decoder $\mathrm{D}_s$, $s\in\{1,2,3,4\}$.
Let $\mathbf{o}^{s}$ and $\mathbf{y}^{s}$ denote the logits and corresponding target at $\mathrm{D}_s$.
We define pairwise exclusion relations among a dataset's foreground labels. Unlike previous topology-informed losses~\cite{shi2023nextou}, which require manually specified relations for a fixed label space, TPUS automatically derives pairwise exclusions from each dataset's native foreground labels. Hard predictions $\mathbf{P}^{s}$ are obtained by taking the class-wise argmax of the softmax probabilities. A $3\!\times\!3\!\times\!3$ kernel is then applied to $\mathbf{P}^{s}$ to identify violations within each voxel's 26-connected neighborhood, yielding a binary critical-voxel mask $\mathbf{V}^{s}$. For batch size $N$ and voxel domain $\Omega_s$, the topological interaction (TI) term applies cross-entropy only at these voxels:
\begin{equation}
    \mathcal{L}^{s}_{\mathrm{T}}=
    \frac{1}{N}\sum\nolimits_{n=1}^{N}\sum\nolimits_{v\in\Omega_s}
    V^{s}_{n,v}\,\operatorname{CE}(\mathbf{o}^{s}_{n,v},\mathbf{y}^{s}_{n,v}).
    \label{eq:ti_loss}
\end{equation}
For a single foreground class, the exclusion set is empty and $\mathcal{L}_{\mathrm{T}}^s=0$; the four UMD foreground classes yield six exclusion pairs. The total loss is formulated as follows: 
\begin{equation}
    \mathcal{L}    =\sum\nolimits_{s=1}^{4}w_s\mathcal{L}^s
    =\sum\nolimits_{s=1}^{4}w_s\left(
    \mathcal{L}^{s}_{\mathrm{Dice}}+\mathcal{L}^{s}_{\mathrm{CE}}+
    \lambda_{\mathrm{T}}\mathcal{L}^{s}_{\mathrm{T}}\right),
    % \quad
    %\lambda_{\mathrm{TI}}=10^{-6},
    \label{eq:overall_loss}
\end{equation}
where $\lambda_{\mathrm{T}}$ is empirically set as $10^{-6}$. 
We use normalized deep-supervision weights $w_s=2^{s-1}/15$ for $\mathrm{D}_s$ ($s\in\{1,2,3,4\}$). Only the full-resolution dynamic output at $\mathrm{D}_4$ is used at inference.

% Declared before the experiments so the double-column float can appear on page 4.
\begin{figure*}[!t]
\setlength{\abovecaptionskip}{-4pt}
\setlength{\belowcaptionskip}{0pt}
\setlength{\abovedisplayskip}{0pt}
\setlength\belowdisplayskip{0pt}
    \centering
    \includegraphics[width=\textwidth]{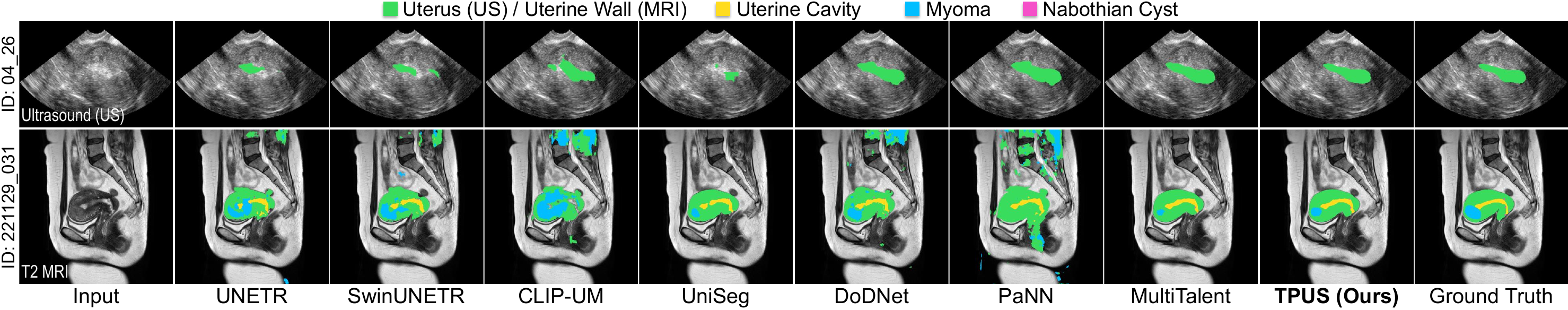}
    \caption{%Qualitative comparison on the ultrasound image from UterUS (top) and MRI from UMD (bottom): input, seven baselines, TPUS, and ground truth.
    Visual comparison of segmentation results on the UterUS (top) and UMD (bottom) datasets. Shown are the input, competing methods, the proposed TPUS, and ground truth. Different colors denote distinct tissues.}
    \label{fig:qualitative_comparison}
\end{figure*}

\section{EXPERIMENTS}
\label{sec:experiments}

\subsection{Data and Implementation}
The UterUS dataset~\cite{bones2024automatic} comprises 3D US volumes with binary uterine-region masks, excluding cases with fibroids, adenomyosis, and other acquired abnormalities. 
The Uterine Myoma MRI Dataset (UMD)~\cite{pan2024umd} provides labels for the \textit{uterine wall}, \textit{uterine cavity}, \textit{myoma}, and \textit{Nabothian cyst} (when present) on T2-weighted MRI of patients with surgically and pathologically confirmed fibroids.
Training/validation/held-out test sets contain 100/10/28 cases for UterUS and 191/21/54 for UMD. Following MultiTalent~\cite{ulrich2023multitalent}, we adopt its nnU-Net-based preprocessing~\cite{isensee2021nnunet} (1-mm isotropic resampling and z-score normalization) and data augmentation. Training uses $192^3$ patches, batch size 1, and 1,000 epochs of 250 iterations with Nesterov SGD (momentum 0.99, weight decay $3\!\times\!10^{-5}$; polynomial decay from $10^{-3}$). All methods share splits, preprocessing, and postprocessing. Dice and average symmetric surface distance (ASSD) are reported as mean$\pm$SD over cases per foreground class, then macro-averaged; ASSD is omitted for empty surfaces.

% Declared mid-method to place the primary results at the top of page 3.
\begin{table}[htbp]
\centering
\caption{%Test 
Segmentation results on UterUS and UMD.}
\label{tab:comp_final}
\footnotesize
\renewcommand{\arraystretch}{0.9} 
\setlength\tabcolsep{0.5pt}
\begin{tabular}{@{}l|cccc@{}}
\toprule
\multirow{2}{*}{Method} & \multicolumn{2}{c}{UterUS} & \multicolumn{2}{c}{UMD} \\ \cmidrule(lr){2-3} \cmidrule(lr){4-5}
 & Dice $\uparrow$ & ASSD (mm) $\downarrow$ & Dice $\uparrow$ & ASSD (mm) $\downarrow$ \\ \midrule
UNETR~\cite{hatamizadeh2022unetr} & \msd{0.792}{0.110} & \msd{0.846$^*$}{0.420} & \msd{0.330}{0.199} & \msd{18.899$^*$}{21.245} \\
SwinUNETR~\cite{hatamizadeh2022swinunetr} & \msd{0.864}{0.071} & \msd{0.593$^*$}{0.304} & \msd{0.347}{0.200} & \msd{15.065$^*$}{20.817} \\
CLIP-UM~\cite{liu2023clipdriven} & \msd{0.861}{0.054} & \msd{0.637$^*$}{0.251} & \msd{0.144}{0.137} & \msd{19.165$^*$}{11.165} \\
UniSeg~\cite{ye2023uniseg} & \msd{0.498}{0.160} & \msd{1.715$^*$}{0.513} & \msd{0.599}{0.242} & \msd{2.732$^*$}{4.737} \\
DoDNet~\cite{zhang2021dodnet} & \msd{0.891}{0.050} & \msd{0.760$^*$}{1.169} & \msd{0.433}{0.197} & \msd{17.143$^*$}{19.048} \\
PaNN~\cite{zhou2019pann} & \msd{0.780}{0.095} & \msd{2.242$^*$}{2.099} & \msd{0.273}{0.158} & \msd{20.466$^*$}{9.162} \\
MultiTalent~\cite{ulrich2023multitalent} & \msd{0.895}{0.051} & \msd{0.448$^*$}{0.162} & \msd{0.680}{0.257} & \msd{1.775$^*$}{3.880} \\
\textbf{TPUS (Ours)} & \bmsd{0.898}{0.050} & \bmsd{0.432}{0.156} & \bmsd{0.693}{0.254} & \bmsd{1.684}{4.680} \\
\bottomrule
\end{tabular}
\end{table}

\subsection{Experimental Results} 
We compare the proposed TPUS with task-specific \textbf{UNETR}~\cite{hatamizadeh2022unetr} and \textbf{SwinUNETR}~\cite{hatamizadeh2022swinunetr}, and five joint-training methods: \textbf{CLIP-UM} with CLIP embeddings~\cite{liu2023clipdriven}, \textbf{UniSeg} with task prompts~\cite{ye2023uniseg}, \textbf{DoDNet} with task-conditioned filters~\cite{zhang2021dodnet}, \textbf{PaNN} with size priors~\cite{zhou2019pann}, and \textbf{MultiTalent} with dataset-specific heads~\cite{ulrich2023multitalent}.
Table~\ref{tab:comp_final} reports results of the eight methods on UterUS and UMD. 
%Table~\ref{tab:comp_final} compares TPUS with two task-specific transformer baselines (\ie, UNETR and Swin UNETR~\cite{hatamizadeh2022unetr,hatamizadeh2022swinunetr}) and five universal or multi-dataset methods. The five joint-training baselines use CLIP semantic label embeddings~\cite{liu2023clipdriven}, learned task prompts~\cite{ye2023uniseg}, task-conditioned dynamic filters~\cite{zhang2021dodnet}, anatomical size priors~\cite{zhou2019pann}, or shared training with dataset-specific heads~\cite{ulrich2023multitalent}. 
Table~\ref{tab:comp_final} shows that TPUS achieves the best Dice and ASSD on UterUS (0.898 and 0.432~mm vs. 0.895 and 0.448~mm for MultiTalent) and UMD (0.693 and 1.684~mm vs. 0.680 and 1.775~mm). 
%We also compare TPUS and MultiTalent using five-fold cross-validation; two-sided paired Wilcoxon signed-rank tests on case-wise ASSD give $p=0.00775$ for UterUS and $p=0.0259$ for UMD. 
Two-sided paired Wilcoxon signed-rank tests comparing TPUS with each competing method  in terms of ASSD demonstrate that TPUS achieves statistically significantly better performance ($p$$<$$0.05$, marked with $^*$). 
%然后呢？是不是需要一句话分析一下我们方法好的原因？

%Figure~\ref{fig:qualitative_comparison} qualitatively compares the visual performance across methods on both datasets. Specifically, on the challenging 3D ultrasound dataset ({UterUS}), competing baselines often suffer from severe target fragmentation or undersegmentation due to acoustic shadowing and low soft-tissue contrast; in contrast, our {TPUS} benefits from the graph-based feature reasoning and topology-informed supervision, producing a remarkably compact, anatomically contiguous, and boundary-complete uterine mask. On the multi-class T2-weighted MRI dataset ({UMD}), several competing methods misclassify or generate noisy, distant false-positive components scattered across pelvic tissues, while TPUS preserves the delicate structural configuration of the uterine wall and cavity. Furthermore, it accurately localizes and delineates complex focal lesions (such as myomas and Nabothian cysts) with sharp boundaries, demonstrating the effectiveness of prompt-conditioned dynamic heads in adapting to heterogeneous label spaces and multi-structure contexts. 
Figure~\ref{fig:qualitative_comparison} qualitatively compares different methods on both datasets. On the challenging 3D US dataset ({UterUS}), competing methods often exhibit target fragmentation or under-segmentation due to acoustic shadowing and low soft-tissue contrast, whereas {TPUS} produces compact, anatomically contiguous, and boundary-complete uterine masks. On the multi-class T2-weighted MRI dataset ({UMD}), TPUS suppresses noisy and distant false positives while preserving the uterine wall and cavity and accurately delineating focal lesions, such as myomas and Nabothian cysts, with sharp boundaries. This highlights TPUS's ability to maintain structural integrity and adapt to heterogeneous multi-structure label spaces.

\if false 
{\color{red}
% 这个Figure分析太简略了，几乎没说什么
Figure~\ref{fig:qualitative_comparison} qualitatively supports these results: TPUS produces a compact, contiguous UterUS mask, whereas several baselines undersegment or fragment the target; on UMD, it preserves the wall--cavity configuration and localizes the myoma without the distant false-positive components observed in several baselines. 
}
\fi

\begin{table}[htbp]
\centering
\caption{Ablation results on all 54 test cases from UMD.}
\label{tab:abla_umd}
% \footnotesize
\small
\renewcommand{\arraystretch}{0.9} 
\setlength\tabcolsep{3pt}
\begin{tabular}{@{}l|cc@{}}
\toprule
~Configuration & Dice $\uparrow$ & ASSD (mm) $\downarrow$\\
\midrule
~Base & \msd{0.677}{0.258} & \msd{1.834}{4.608} \\
~Base + Prompt & \msd{0.678}{0.253} & \msd{1.735}{4.128} \\
~Base + Prompt + TopoBlock & \msd{0.680}{0.259} & \msd{1.718}{3.383} \\
~Base + Prompt + DynConv & \msd{0.684}{0.255} & \msd{1.606}{4.787} \\
~Base + TI & \msd{0.686}{0.255} & \bmsd{1.536}{3.237} \\
~\textbf{TPUS (Ours)} & \bmsd{0.693}{0.254} & \msd{1.684}{4.680} \\
\bottomrule
\end{tabular}
\end{table}

\subsection{Ablation Study}
%Table~\ref{tab:abla_umd} evaluates the contribution of each key component in TPUS under a matched protocol on the UMD dataset. Here, ``Base'' denotes the MultiTalent multi-dataset architecture retrained under our experimental protocol. We progressively incorporate our proposed modules: (1) adding the prompt-conditioning mechanism (\textbf{Base + Prompt}) slightly improves the Dice score; (2) further integrating the graph neural network (\textbf{Base + Prompt + GNN}) enhances multi-scale long-range feature reasoning; (3) incorporating the dynamic convolution head (\textbf{Base + Prompt + DynConv}) better adapts to dataset-specific outputs; and (4) introducing the topology-informed supervision alone (\textbf{Base + TI}) effectively reduces boundary errors. Finally, the full \textbf{TPUS} framework combines all modules, achieving the best overall performance and validating the necessity and complementarity of each component.
Table~\ref{tab:abla_umd} evaluates each TPUS component under a matched protocol on UMD. 
``Base'' denotes the MultiTalent model retrained under our protocol. 
Adding prompt conditioning ({Base+Prompt}) yields a slight Dice gain, while TopoBlock integration ({Base+Prompt+TopoBlock}) improves multi-scale long-range reasoning, and dynamic convolution ({Base+Prompt+DynConv}) enhances dataset-specific adaptation. 
Topology-informed deep supervision ({Base+TI}) reduces boundary errors. 
Combining all components in TPUS achieves the highest Dice, demonstrating their complementary contributions.

\if false 
{\color{cyan}CHOOSE ONE VERSION. CLEARER VERSION}

Table~\ref{tab:abla_umd} evaluates TPUS components under the matched protocol, with ``Base'' denoting the MultiTalent architecture retrained under the ablation protocol. 
% 请问：写成这样读者能看懂不？
Prompt alone increases Dice from 0.677 to 0.678; adding TopoBlock or DynConv to Prompt yields 0.680 and 0.684, respectively, while TI alone yields 0.686. The full TPUS achieves the highest Dice, whereas TI alone yields the lowest ASSD (1.536~mm), indicating that improvements in overlap and boundary distance need not coincide.
\fi

\subsection{Influence of Prompt-Encoding Strategy}

We evaluate the impact of different prompt-encoding strategies within the TPUS framework, comparing our default CLIP text encoder against a prompt-free FiLM approach~\cite{perez2018film}, the UniSeg-style task prompt~\cite{ye2023uniseg}, and a randomly initialized learnable embedding. 
As reported in Table~\ref{tab_promptEncoding}, 
removing explicit prompt conditioning or using alternative schemes slightly degrades segmentation accuracy. 
The prompt-free FiLM and UniSeg-style prompt variants achieve Dice score of 0.893, while the random embedding strategy reaches 0.896. 
The proposed pretrained CLIP text encoder achieves the highest Dice and lowest ASSD scores ($p$$<$$0.05$ via two-sided paired Wilcoxon signed-rank tests). 
\if false 
These results imply that semantically rich CLIP-derived prompts %{\color{red}better} 
more effective guidance for %guide 
segmentation. 
\fi 
These results suggest that semantically rich CLIP-derived prompts used in TPUS provide more effective guidance for segmentation.

\begin{table}[!tbp]
\centering
\caption{TPUS with different prompt-encoding methods on UterUS.}
\label{tab_promptEncoding}
\footnotesize
\renewcommand{\arraystretch}{0.9} 
\setlength\tabcolsep{3pt}
\begin{tabular}{@{}l|ccc@{}}
\toprule
~Method & Dice $\uparrow$ & ASSD (mm) $\downarrow$ & $p$-value \\
\midrule
~Prompt-free FiLM~\cite{perez2018film} & \msd{0.893}{0.048} & \msd{0.456}{0.154} & $1.72\!\times\!10^{-8}$ \\
~UniSeg-style Prompt~\cite{ye2023uniseg} & \msd{0.893}{0.050} & \msd{0.463}{0.149} & $1.53\!\times\!10^{-4}$ \\
~Random Embedding & \msd{0.896}{0.047} & \msd{0.448}{0.148} & ${1.17\!\times\!10^{-3}}$ \\
~\textbf{CLIP Text Encoder (Ours)}  & \bmsd{0.898}{0.050} & \bmsd{0.432}{0.156} & -- \\
\bottomrule
\end{tabular}
\end{table}

\section{CONCLUSION}
\label{sec:conclusion}
{This work presents TPUS, a unified framework for multi-structure uterine segmentation across ultrasound and MRI. }
TPUS integrates a graph-based multi-dataset backbone comprising modality-specific stems and a modality-shared graph-based encoder-decoder, prompt-conditioned dynamic prediction, and topology-informed deep supervision for heterogeneous uterine ultrasound and MRI datasets. It achieves the highest Dice and lowest ASSD among the compared methods on both held-out test sets. 
A current limitation is that evaluation is restricted to one ultrasound and one MRI dataset; future work will assess scalability and generalizability by integrating additional datasets with diverse acquisition protocols and label spaces.

\section{COMPLIANCE WITH ETHICAL STANDARDS}
%Only publicly available, 
De-identified UterUS and UMD data were analyzed in this work. The UMD dataset had Beijing Shijitan Hospital ethics approval  (sjtkyll-lx-2022(1)) 
and waived consent; %~\cite{pan2024umd}; 
this secondary analysis required no additional approval.

\section{ACKNOWLEDGMENTS}
This work was supported in part by NIH grants AG073297, AG082938, EB035160, and NS134849. The authors declare no relevant financial or non-financial conflicts of interest.

%\vfill
%\pagebreak


% -------------------------------------------------------------------------
\balance
% \small
%\footnotesize
\bibliographystyle{IEEEbib}
\begingroup
\interlinepenalty=10000
\let\v\originalCaronAccent
\bibliography{ISBI2027}
\endgroup

\end{document}

%% file: defs.tex
\def\v{{\bf v}}

\def\0{{\bf 0}}
\def\1{{\bf 1}}